\documentclass[twocolumn,fleqn,10pt]{wlscirep}
\usepackage[utf8]{inputenc}
\usepackage[T1]{fontenc}
\usepackage{algorithm}
\usepackage{algpseudocode}
\usepackage{amsmath,amssymb,amsfonts}
\usepackage{booktabs}
\usepackage{multirow}
\usepackage{booktabs}
\usepackage{multirow}
\usepackage{algorithmicx}
\usepackage{algpseudocode}
\title{Generative artificial intelligence-enabled dynamic detection of nicotine-related circuits}

\author[1,2]{Changwei Gong}
\author[1,2]{Changhong Jing}
\author[1]{Ye Li}
\author[1]{Xianan Liu}
\author[1]{Zuxin Chen}
\author[1,*]{Shuqiang Wang}
\affil[1]{Shenzhen Institutes of Advanced Technology, Chinese Academy of Sciences, Shenzhen, 518060, China}
\affil[2]{University of Chinese Academy of Sciences, Beijing, 100049, China}

\affil[*]{sq.wang@siat.ac.cn}

\begin{abstract}
The identification of addiction-related circuits is critical for explaining addiction processes and developing addiction treatments. And models of functional addiction circuits developed from functional imaging are an effective tool for discovering and verifying addiction circuits. However, analyzing functional imaging data of addiction and detecting functional addiction circuits still have challenges. We have developed a data-driven and end-to-end generative artificial intelligence(AI) framework to address these difficulties. The framework integrates dynamic brain network modeling and novel network architecture networks architecture, including temporal graph Transformer and contrastive learning modules. A complete workflow is formed by our generative AI framework:  the functional imaging data, from neurobiological experiments, and computational modeling, to end-to-end neural networks, is transformed into dynamic nicotine addiction-related circuits. It enables the detection of addiction-related brain circuits with dynamic properties and reveals the underlying mechanisms of addiction. 
\end{abstract}
\begin{document}

\flushbottom
\maketitle
\thispagestyle{empty}

\clearpage
\section*{Introduction}

Nicotine addiction caused by smoking is the most common drug addiction behavior of human beings worldwide.\cite{le2022tobacco} Its hazards are cardiovascular disorders, respiratory diseases, cancers, and other diseases caused by long-term smoking, which has become a global problem threatening human health. Addiction is mainly due to the dependence of addictive substance, and nicotine is considered one of the most addictive drugs. The essential cause of addiction is related to the dramatic impairment of brain circuits.\cite{koob2016neurobiology} Many studies have been recorded to investigate the changes in the brain of addicts and conceptualize addiction as a functional brain disease.\cite{kim2017molecular} However, nicotine dependence is difficult to treat, and the biological mechanisms are not fully clarified.\cite{luscher2016emergence} 

Neuroimaging have revealed neurochemical and functional changes in the brains of addicted individuals\cite{fMRI}, providing new insights into the mechanisms of addiction. Functional magnetic resonance (fMRI) has been used to study nicotine dependence's neural basis and develop smoking cessation strategies\cite{bifone2011functional}. Because acute drug administration's neurological and behavioral effects tend to be of short duration, fMRI is better than PET for identifying brain circuits activated during drug administration. It is critical to obscure temporal patterns in the short term. Such dynamic changes can be detected by functional magnetic resonance imaging (fMRI), which can reflect averages over shorter periods (e.g., <5 seconds). Functional magnetic resonance imaging (fMRI) studies have revealed complex dynamic patterns of brain changes during drug intoxication, with different temporal patterns when some regions are activated and others restrained.\cite{volkow2003addicted} In some regions, the time series of these dynamic patterns of change parallel the patterns of behavioral effects of drugs of abuse\cite{vaquero2017cocaine,menossi2013neural}.

Animal models and human imaging studies have made it possible to investigate the brain system and neural circuits disrupted by drug addiction.\cite{bruijnzeel2015acute,claus2013association}.Multiple interconnected neural circuits, such as heightened incentive salience and habit development, reward deficiencies and stress overload, and poor executive function, are existing instances.\cite{ikemoto2015basal,nestler2019molecular}. Brain imaging studies focus on specific brain circuits producing a compound addiction cycle, including intoxication, withdrawal, and craving\cite{schultz2011potential}. However, brain imaging investigations of this component of the addiction cycle and relevant neural circuits engaged in the motivational elements of drug withdrawal throughout the withdrawal phase, including acute administration of drugs following abstinence, have rarely been explored. Therefore, the investigation of abnormalities in brain imaging of the circuits triggered by nicotine addiction and the detection and analysis of these addiction-related circuits under the specific phase mentioned above are the primary research goals of this work.

Meanwhile, generative artificial intelligence(AI)\cite{castelli2022generative} has recently been applied to several emerging medical and neuroscience fields.  With the advent of related theories and techniques, we can train much larger models than ever and perform much more effective. This has greatly improved the performance of many tasks, and the architecture of learning models is steadily improving. Researchers now better understand how to design effective neural networks, and many new architectures have been proposed in recent years. Examples include Generative Adversarial Networks (GAN)\cite{GAN} and Transformer\cite{vaswani2017attention}. While the conclusions of generative AI will vary depending on the application or domain under consideration, the potential of generative AI can create new, previously unimaginable things. The capabilities of generative AI can help us better understand and model the complex functions of complex systems, such as the brain. Towards the complex system in neuroscience and its challenges, generative AI will be able to make a big difference. The collaboration between generative AI and brain imaging will provide new impetus and approaches to explore complex brain function and addiction mechanisms.

Here, we perform nicotine withdrawal and acute drug administration experiments in a rat animal model to simulate smoking addiction, obtain functional MRI data, and analyze drug addiction imaging experiments with generative intelligence-enabled detection of nicotine addiction-related circuits in rats. The proposed process is the first to adopt a data-driven, end-to-end approach, supported by biologic imaging experiments, to explain the underlying abnormal mechanisms in addicted individuals, incorporating the comprehensive knowledge of neurosciences. In terms of the efforts to achieve our goals, this work contributes to the three challenges noted below: (1)Addiction rat model with high intensity fMRI, (2) Generative intelligence for brain network, and (3)Interpretable addiction-related circuits.

Firstly, to better simulate the addictive state of smokers in animal models and to more accurately observe their physiopathological changes, we established a rat or mouse model of nicotine self-administration. This experimental model better simulates the spontaneous giving behavior of smokers than other existing addiction models. We used a Bruker $7$ Tesla scanner to perform the functional magnetic resonance imaging experiment on rats. Validation of brain loops obtained from fMRI whole brain functional imaging analysis will be performed to find clues and potential biomarkers for interventions in smoking addiction.

Next, to specifically study the connectivity between brain regions, we transform the fMRI data from rat addiction experiments into network data to analyze addiction-related circuits in the form of nodes and connections. We employ graph representation learning to capture the latent non-Euclidean graph features in rat addiction brain networks. To address the problem of the small sample size of bio-experimental data, a generative AI with contrast learning is designed to process small sample bio-imaging data. Homo-modality rat data are added for pre-training, and several techniques are used to mitigate the overfitting of learning models trained on a small sample of data. It can avoid the invalidity of detection of nicotine addiction-related circuits.

Thirdly, to capture changes in instantaneous addiction-related circuits during acute drug administration in addiction withdrawal, we designed the architecture of temporal networks in the model specifically for dynamic brain networks out of consideration for the design of addiction experiments in rats to represent the temporal alterations in addiction brain circuits by fMRI signals. It would yield our expected target results characterizing the changes in pharmacology upon acute nicotine injection. And we construct an expert system from relevant references in neurobiology and neuroimaging to interpret the results detected by artificial intelligence through prior validation in neuroscience to uncover new brain regions with unknown neural circuit connections in nicotine addiction issues. It also forms a platform for the development of experimental design and imaging studies on various drug addictions.

\section*{Results}

\begin{figure*}[htbp]
\centering
\includegraphics[width=140mm]{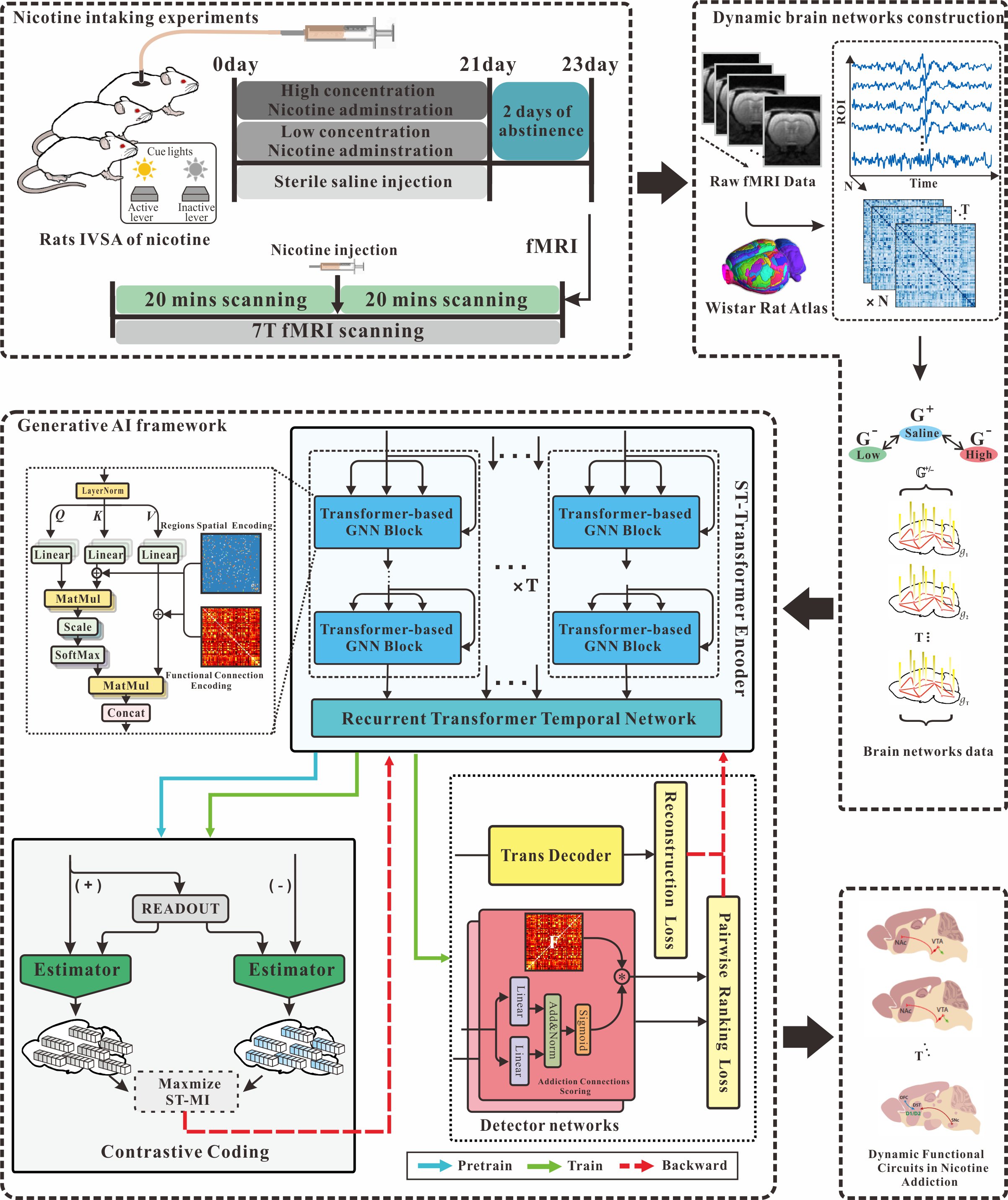}
\caption{RACD platform.}
\label{fig1}
\end{figure*}

\begin{figure*}[!t]
\centering
\includegraphics[width=150mm]{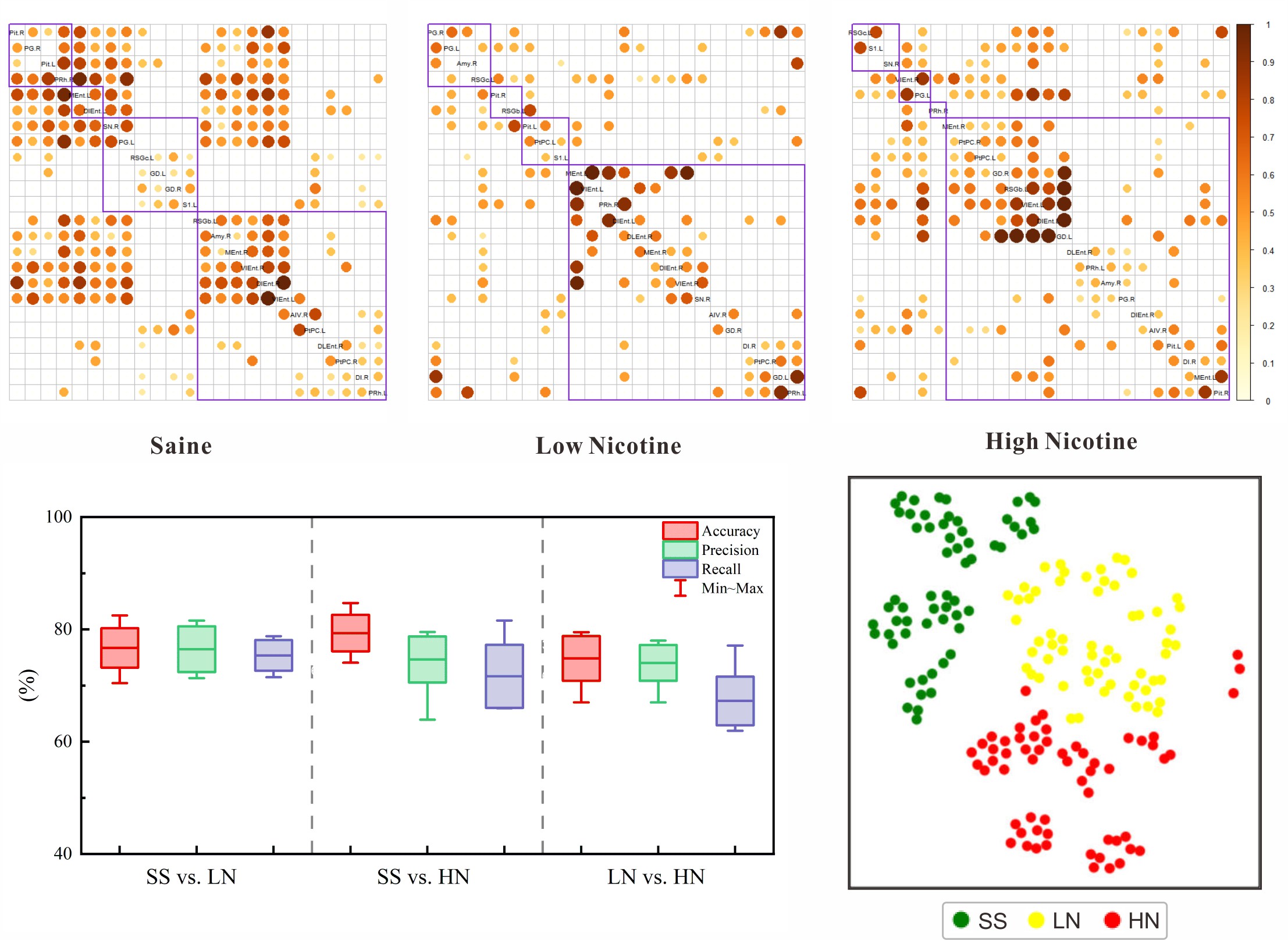}
\caption{AI-enabled generative model.}
\label{fig2}
\end{figure*}

\subsection*{RACD platform.}We developed Rat Addiction-related Circuits Detection(RACD)(Fig.1), a hybrid pipeline from designed experiment with rat models to end-to-end artificial intelligence framework for exploring abnormal circuits in rat nicotine-addiction experiments. Pharmacological brain imaging experiments on addiction in rats were designed and implemented at the beginning of the process, and then we used the raw fMRI data to create a rat addiction brain network dataset through a series of brain network construction techniques. To effectively uncover these underlying and dynamic patterns, we bridged generative intelligence methods to extract such dynamic topological representations and detect dynamic addiction-related circuits from them.
In the experimental design, to simulate withdrawal from nicotine addiction, rats were injected with nicotine for 21 days to simulate long-term smoking addiction, and then absent for two days. Moreover, in order to capture the pharmacological fMRI of withdrawal and reintroduction of nicotine to activate the addiction circuit, we administered the corresponding dose of nicotine to the rats at the middle of the fMRI scanning to obtain brain images with alterations in the addiction neural circuit.

After that, the fMRIs of the three groups of rats were preprocessed to get the dynamic functional connectivity (FC) of the brain. To fix the stratification confusion caused by the interleaved sequence of odd layers during the data set; head shift correction, align the data of all time points of each object with the data of the first time point to get rid of head shift artifacts and reduce the effect of head shaking during scanning; registering all of the objects' data in the same space. Then, we divided the fMRI image space into 150 brain regions by Wistar brain atlas\cite{brain_atlas}; and averaged the time series of all voxels in a specific ROI with the time series of all voxels in a specific ROI to get the time series of all brain regions for each object. Finally, we calculated the correlation coefficients of each other regions as the connectivity information of brain networks.

The generative-constrastive learning framework consists of spatiotemporal graph transformer(SGT) as an encoder, spatiotemporal representation contrasting, and addiction-related functional circuits identification. The encoder(E) takes the dynamic brain networks as input to perform topological information propagation between brain regions at different time steps ,each brain region can aggragate features from its neighboring regions and each timestamp can merge the temporal information at the neighboring time slice. Given the latent representations learned from encoder, spatiotemporal representation contrasting employs a estimator(E) to represent the probability scores assigned to positive pairs and negative pairs and maximizes mutual information between current graph representation and the pooled spatiotemporal global representation.Its objective is pretrainning and trainning the encoder to obtain useful representations to detect addiction related circuits between nicotine groups and saline group. Then a scoring network is adopted in addiction related functional circuits identification to compute the score for each brain network connection which represent the possibilities of addiciton related circuits. Furthermore, the loss function incorporating by the InfoMax loss and the pairwise ranking loss encourages to learning useful representation and discover effective addiction related circuits.

\begin{figure*}[h]
\centering
\includegraphics[width=125mm]{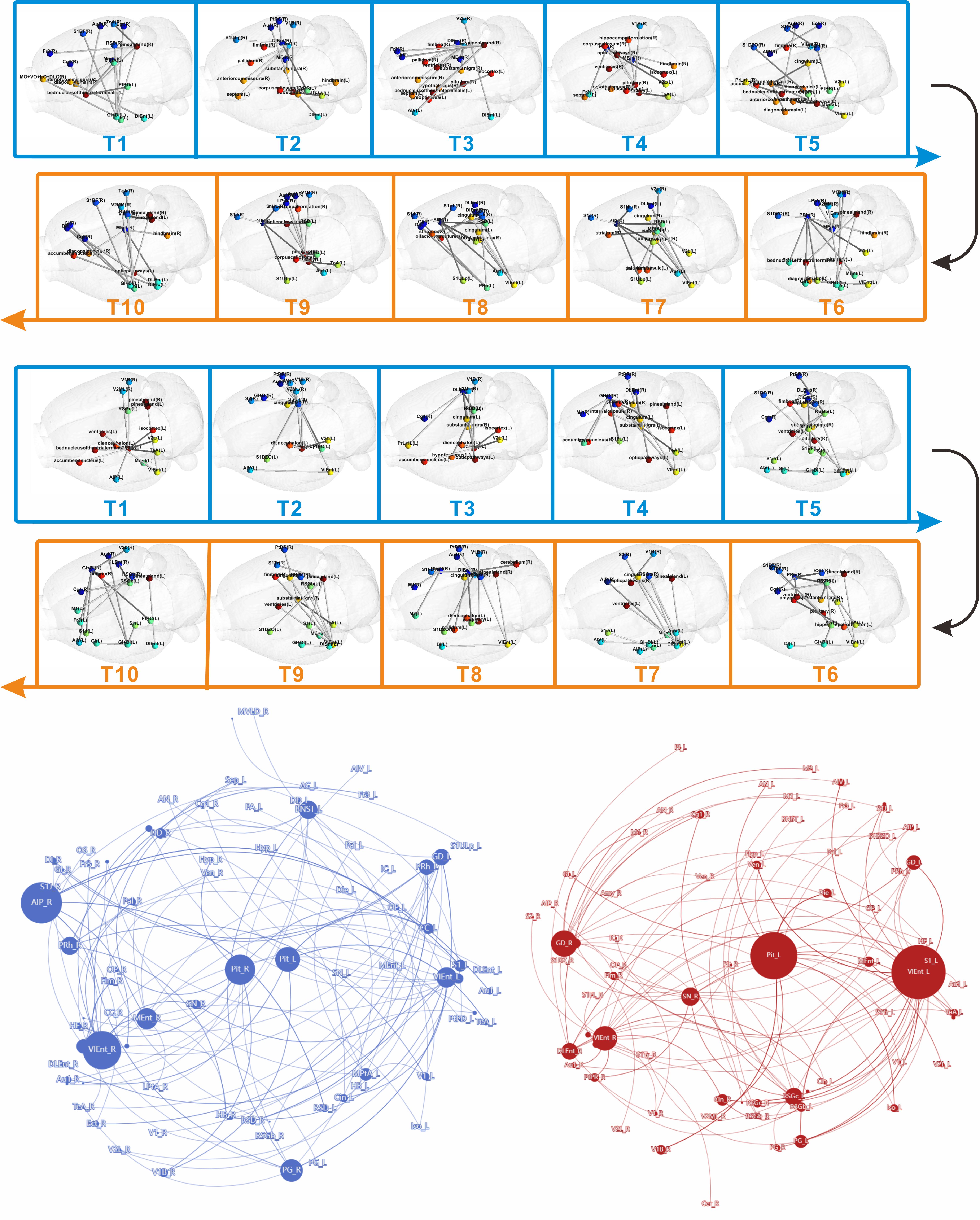}
\caption{Addiction-related circuits detection.}
\label{fig3}
\end{figure*}

\subsection*{AI-enabled generative model.}To take the most advantages of generative models in dealing with small samples and to investigate the different brain networks generated under the nicotine injection, the backbone of the proposed end-to-end generative learning framework is implemented. Labeled rat brain network data is fed to the graph autoencoder network, and this part of generative model can learn the distribution of addiction-irrelevant normal rat brain networks. Especially the pre-training data of normal control brain network is provided to this part of the generative model can effectively mitigate the small sample problem of rat addiction. Here, the coding of the different distributions of addiction-irrelevant and addiction-relevant representations embedded in the hidden space is important because it can determine the quality of the detected addiction circuits. To evaluate this encoding information and the generated model, we carried out experiments with the classification task in the field of AI, which can be considered that good classification results will be more capable of distinguishing addiction feature patterns and contribute to the detection of circuits.

\begin{figure*}[ht]
\centering
\includegraphics[width=145mm]{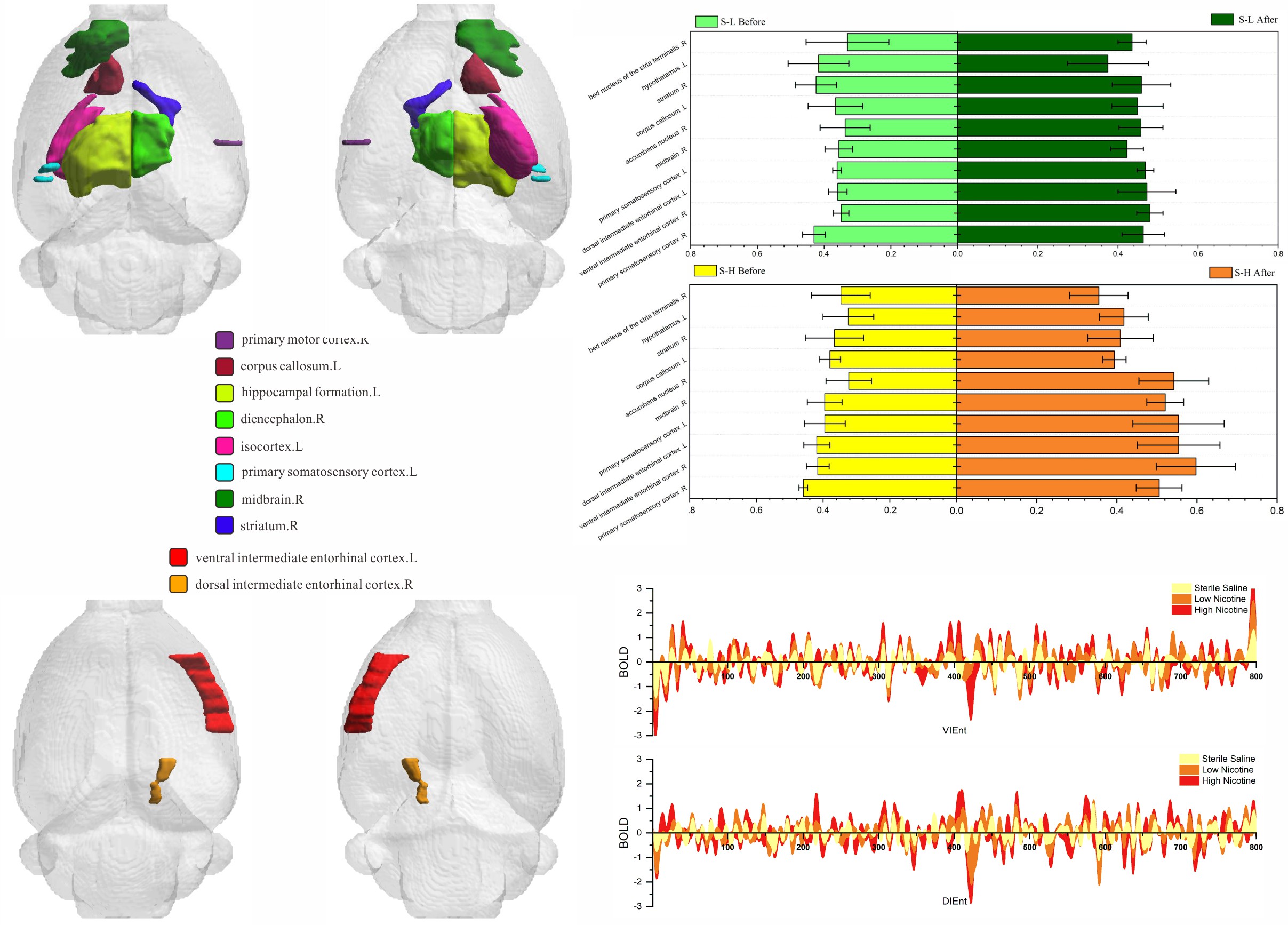}
\caption{Addiction-relevant regions exploration.}
\label{fig4}
\end{figure*}

\begin{figure*}[ht]
\centering
\includegraphics[width=130mm]{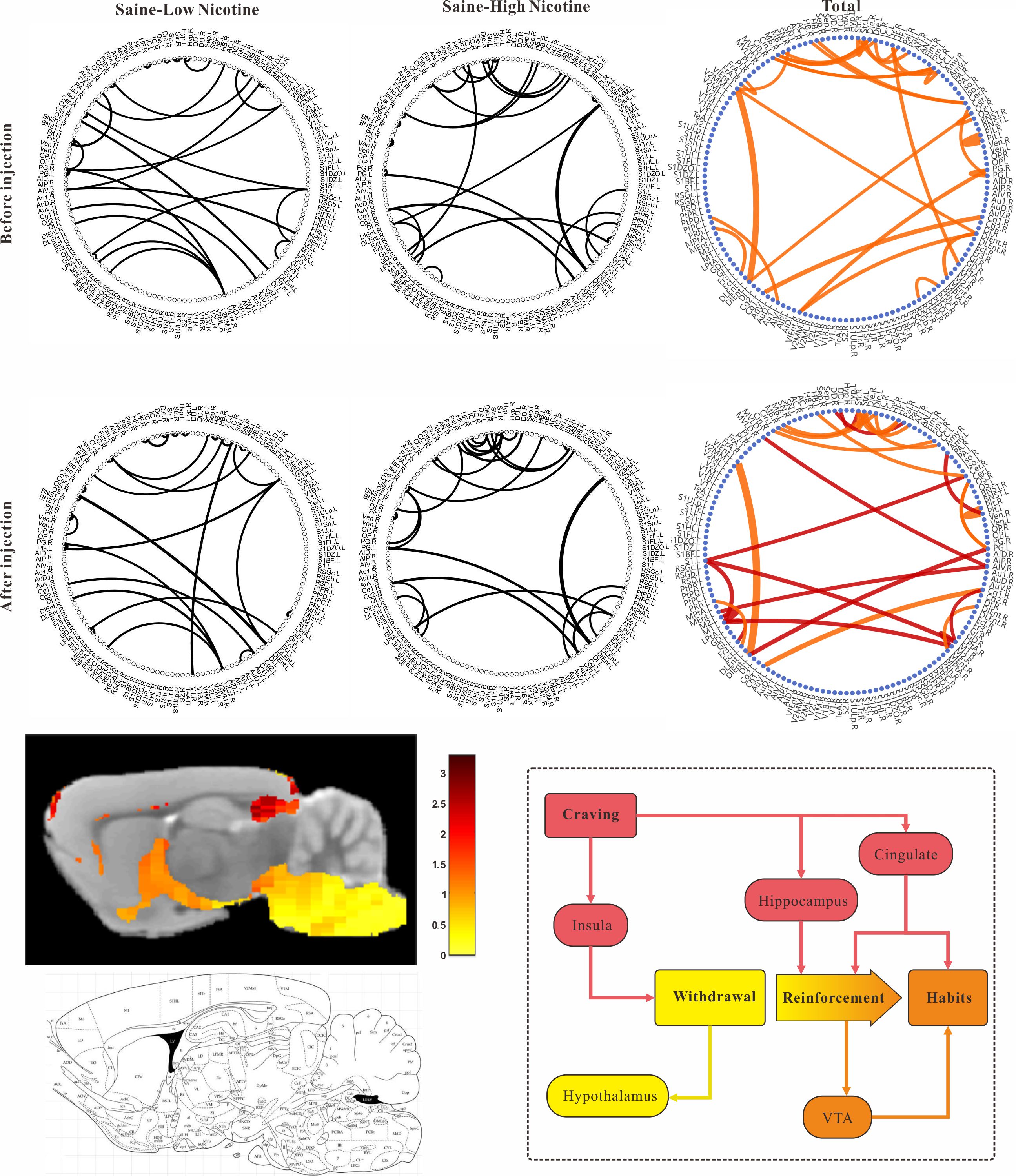}
\caption{Discussion.}
\label{fig5}
\end{figure*}

In this generative AI framework, the functional connections of the three dynamic brain networks corresponding to the three types of inputs can be generated, i.e., the functional connection matrices of the saline group, the low nicotine group and the high nicotine group are reconstructed. In the generated matrices, we select and preserve the connections of important and determinative brain regions as shown in fig2a, fig2b, fig2c. The accuracy, precision and recall are considered as metrics to demonstrate the effectiveness of the generated brain networks for addiction-related features, and we visualize these features by the t-SNE method, as shown in fig2e. It is found that the selected features of different groups of brain regions are differentiable and the generated connections in the proposed framework are effective and beneficial for the detection of addiction from the artificial intelligence point of view.

\subsection*{Addiction-related circuits detection.}After the effective reconstructed brain connections has been achieved, paired brain connections are input to a detector network of addiction connection scores, and with the assistance of contrast learning methods, two groups of addiction-related circuits scores are obtained through training, the saline-low nicotine (S-L) group and the saline-high nicotine (S-H) group, respectively, and we expect that this paired design will achieve the goal of finding out the spatiotemporal nicotine effects of differential changes. It is worth noting that we divided the time series of the entire brain network into 10 time steps because of the temporal resolution of fMRI, which is a division strategy that captures as much useful temporal variation as possible. Moreover, we treated the top 1 percent of scored connections as addiction-related abnormal circuits to obtain the most precise and interpretable detection results.

Therefore, the results of the single sample output are shown in fig3a and fig3b, corresponding to the S-L and S-H groups. With the first 5 time steps boxed in blue, it indicates the difference without acute injection between simulated addiction withdrawal and normal control, and the rest 5 time steps boxed in orange, indicating after acute injection, the difference between simulated addiction relapse and normal control. It can be observed from different times that although the addiction-related connections are changing at each time step, the connections are scattered by a number of central nodes(hubs) of brain regions and make the brain nodes interconnected through these hubs. Some of these important centers and the connections between them will remain present at specific time steps. When comparing different doses of nicotine in paired groups, the same circuits are partially present at similar or the same time slices, suggesting that the addiction-related brain circuits created by different doses of nicotine injection are similar to some extent, but on the other view, the different addiction-related circuits suggest that different doses may also correspond to different addiction mechanisms. Ignoring temporality and individual specificity, we count the outcomes of all samples in the test set to show all the circuits present in the S-L and S-H groups under the whole time series, and show the size of brain nodes according to the cumulative value of the score weights of the circuits, as shown in fig3c and fig3d.

\subsection*{Addiction-relevant regions exploration.}Under this study, we created a new SBT model based on functional magnetic resonance imaging (fMRI) of the rat brain in order to pinpoint brain areas and brain connections linked with nicotine addiction. Furthermore, the suggested SRC module creates contrastive objects between the positive and negative samples, using adversarial learning and feature matching to further enhance the overall performance of the model as a whole. A huge number of tests are carried out in order to validate the efficiency of the model before it is implemented. A comparison was made between the findings generated by the model and those described in the existing literature. Furthermore, the findings collected will be reviewed and confirmed by qualified medical professionals. As a result, such a framework has a wide range of applications and has the potential to be very useful in the identification of neuroimaging biomarkers.

Observe the differences in brain connection between major brain regions while doing the High vs. Low vs. Saline challenge, and highlight areas with enhanced brain connectivity and areas with reduced brain connectivity. These addiction-related connections are the same ones that have been discovered in prior studies. The connection patterns are consistent, and the brain connectivity patterns that have been discovered to be consistent are presented in a figure below.The section measures the variations in the number of connections between major brain areas over time, as defined by several label categories. Furthermore, the modifications observed indicate the metabolic compensatory mechanism of the nicotine-dependent brain, which is consistent with the findings of earlier studies.

This section focuses on the most striking differences in brain connectivity found in the three groups of experiments. Fig4a visualized the TOP10 brain regions with the most obvious differences between high concentration and low concentration, high concentration and normal saline, and low concentration and normal saline, respectively.

\section*{Discussion}

Most of the top-ranked addiction-related brain connectivity obtained by the experiment can be verified in the previous literature. For example, the brain connectivity that differed most between the high and normal saline groups included:
(Hyp\_R, Pit\_R)\cite{brain_region_1}, 
(Cin\_L, Pit\_L)\cite{brain_region_2},
(S2\_R, Cer\_L)\cite{brain_region_3}, 
(S1ULp\_R, Cer\_R)\cite{brain_region_4}, 
(S2\_R, Cer\_R)\cite{brain_region_5}, 
(RSGc\_L, PG\_L)\cite{brain_region_6}, 
(Cer\_L, GI\_r)\cite{brain_region_7}, 
(GI\_R, Cer\_R)\cite{brain_region_8}, 
(PG\_R, RSD\_L),
(Cer\_L, S1BF\_R), 
(Cer\_L, S1DZ\_R), 
(RSD\_R, PG\_R), 
(BNST\_L, SN\_L), 
(S1DZ\_R, Cer\_R), 
(GD\_R, Cer\_L), 
(Cer\_R, S1FL\_R), 
(RSD\_L, PG\_L), 
(Cer\_R, S1BF\_R), 
(Ect\_L, S1FL\_L).
The brain connectivity that differed most between the low and saline groups included:
(PG\_L, Pal\_L), 
(RSGc\_L, PG\_L), 
(Pit\_R, OP\_R), 
(Pit\_R, Cin\_L), 
(MPtA\_L, Pit\_R), 
(Cin\_R, Pit\_L), 
(Pit\_L, MEnt\_L), 
(S1ULp\_R, Cer\_L), 
(Pit\_R, Hyp\_L), 
(Pit\_R, MEnt\_L), 
(DD\_L, Pit\_R), 
(AC\_R, Pit\_R), 
(MEnt\_L, PtPC\_L), 
(Pit\_R, IC\_R), 
(AN\_L, MEnt\_L), 
(Pit\_L, MPtA\_L), 
(Cg2\_R, DD\_R), 
(Cin\_R, Pit\_R), 
(S1ULp\_R, Cer\_R), 
(Pit\_L, Cg2\_R). 
The brain connectivity that differed most between the high- and low-concentration groups included:
(PG\_R, RSGc\_L), 
(S1ULp\_R, Cer\_L), 
(PG\_L, Die\_L), 
(HF\_L, PG\_L), 
(PG\_L, Fim\_L), 
(Str\_L, PG\_L), 
(PG\_L, IC\_L), 
(PG\_L, Cg1\_R), 
(V2L\_L, PG\_L), 
(PG\_L, V1B\_L), 
(CC\_L, PG\_L), 
(V2MM\_R, PG\_L), 
(PG\_L, PA\_L), 
(Amy\_L, PG\_L), 
(MB\_L, PG\_L), 
(Ven\_L, PG\_L), 
(PG\_L, GI\_L), 
(Iso\_L, PG\_L), 
(Cg1\_L, PG\_L), 
(Ect\_L, PG\_L). 
These experimental results verify the effectiveness of the model proposed in this paper.

This section focuses on the brain areas that showed the most noticeable variations between the two paired sets of tests. Fig5a visualized the TOP30 brain circuits with the most obvious differences between high concentration and low concentration, high concentration and sterility saline, and low concentration and sterility saline.

The brain regions that differed most between the high and low-concentration groups included: PG\_R, Ect\_L, SN\_L, PG\_L, Cer\_R, Cer\_L, AC\_L, Pit\_R, RSGc\_L, Pit\_L, RSGb\_L, RSGc\_R, DD\_R, BNST\_L, Pal\_L, DD\_L, Cg2\_R, Amy\_R, PA\_L, RSGb\_R.
The brain regions that differed most between the high and saline groups included: Pituitary\_R, MEnt\_L, Pit\_L, DD\_R, DD\_L, AC\_L, BNST\_L, MEnt\_R, SN\_L, AC\_R, PA\_L, PA\_R, AIV\_L, AIP\_L, AN\_L, Cer\_R, Cer\_L, Cg2\_R, OP\_L, BNST\_R.
The brain regions that differed most between the low and saline groups included:PG\_L, PG\_R, DIEnt\_L, GD\_R, MEnt\_L, S1\_L, PtPC\_L, S1Sh\_L, Pit\_R, RSGb\_L, MPtA\_L, Pit\_L, RSGc\_L, S1Sh\_R, V1B\_L, RSGc\_R, DLEnt\_R, DD\_L, RSGb\_R, PA\_L.These experimental results verify the effectiveness of the model proposed in this paper.

Between both model's top-ranked nicotine addiction-related brain areas output, some are well associated with addiction, while others are not frequently seen in the current literature or in physicians' cognition. The goal of the approach suggested in this study is to identify uncommon yet highly weighted brain areas. This region of the brain may be associated with nicotine addiction, however this has not been proven. Machine learning approaches will be used to confirm these newly found brain areas using biological rat nicotine addiction tests. These experimental findings may contribute to our understanding of the unique mechanism behind nicotine addiction.

\section*{Methods}

The proposed addiction circuits identification framework,which is based on the novel transformer and the spatiotemporal graph contrastive learning, is showed in \ref{fig1}.In this section, we first introduce the notation and problem setting, and then give an overview of proposed SGCL, last describe the structure and mechanism of the main components in details.

Derived from fMRI preprocessing, the  dynamic rat brain networks $\mathbb{G}$ is a kind of time-series graph structure data takes the form of${\{G^t\} }^T_{t=1}$,where each $G^t={(V^t,E^t,X^t)}$ represents the single static functional brain networks at timestamp $t$, $T$ is the maximum timestamp. $V_t \in \mathbb{R}^{M}$ denote the set of brain regions at time step $t$,and $E_t\in\mathbb{R}^{M\times M}$  is the corresponding brain connection set. An edge $e=(i,j,\omega) \in E^t $ denotes the brian connection between the i-th brain region and the j-th brain region at time $t$  and its weight is $\omega$, $X^t={\{x_i^t \}}_1^n \in\mathbb{R}^{N}$ consists of brain regions' fMRI time series, where $x_i^t$ is the vector of time series associated with node $v_i$ at time step $t$.

From the experimentally obtained dynamic functional brain networks of three different groups of rats: $\mathbb{G}_S$ in the saline-injected group (nicotine non-addicted), $\mathbb{G}_L$ in the low-dose nicotine-injected group (nicotine mildly addicted) and $\mathbb{G}_H$ in the high-dose group (nicotine severely addicted), the goal of this paper is to detect dynamic anomaly connections from three groups' dynamic brain networks and finally identify addiction related functional circuits from these connections.

The SGCL consists of spatiotemporal graph transformer(SGT) as an encoder, spatiotemporal representation contrasting, and addiction-related functional circuits identification. The encoder(E) takes the dynamic brain networks  as input to perform topological information propagation between brain regions at different time steps ,each brain region can aggragate features from its neighboring regions and each timestamp can merge the temporal information at the neighboring time slice. Given the latent representations learned from encoder, spatio-temporal representation contrasting employs a estimator($\mathcal{E}$) to represent the probability scores assigned to positive pairs and negative pairs and maximizes mutual information between current graph representation and the pooled spatiotemporal global representation.Its objective is pretrainning and trainning the encoder to obtain useful representations to detect addiction related circuits between nicotine groups and saline group.  Then a scoring network is adopted in addiction related functional circuits identification to compute the score for each brain network connection which represent the possibilities of addiciton related circuits.Furthermore, the loss function incorporating by the InfoNCE loss and the pairwise ranking loss encourages to learning useful representation and discover effective addiction related circuits.

\bibliography{sample}

\section*{Acknowledgements (not compulsory)}

Acknowledgements should be brief, and should not include thanks to anonymous referees and editors, or effusive comments. Grant or contribution numbers may be acknowledged.

\section*{Author contributions statement}

Must include all authors, identified by initials, for example:
A.A. conceived the experiment(s),  A.A. and B.A. conducted the experiment(s), C.A. and D.A. analysed the results.  All authors reviewed the manuscript. 

\section*{Additional information}

To include, in this order: \textbf{Accession codes} (where applicable); \textbf{Competing interests} (mandatory statement). 

The corresponding author is responsible for submitting a \href{http://www.nature.com/srep/policies/index.html#competing}{competing interests statement} on behalf of all authors of the paper. This statement must be included in the submitted article file.

\end{document}